\documentclass[a4paper,conference]{IEEEtran}

\usepackage[pdftex]{graphicx}
\graphicspath{{./figs/}}
\DeclareGraphicsExtensions{.pdf,.jpeg,.png}
\usepackage{amsmath}
\usepackage{textcomp}

\begin{document}

\title{Learning Knowledge-Rich Sequential Model for Planar Homography Estimation in Aerial Video}

\author{
\IEEEauthorblockN{Pu Li and Xiaobai Liu}
\IEEEauthorblockA{San Diego State University, 
San Diego, California 92182\\
Emails: \{pli5270, xiaobai.liu\}@sdsu.edu}}

\maketitle

\begin{abstract}
This paper presents an unsupervised approach that 
leverages raw aerial videos to learn to estimate planar homographic transformation between consecutive video frames. Previous learning-based estimators work on pairs of images to estimate their planar homographic transformations but suffer from severe over-fitting issues, especially when applying over aerial videos. To address this concern, we develop a sequential estimator that 
directly processes a sequence of video frames and estimates their pairwise planar homographic transformations in batches. We also incorporate a set of spatial-temporal knowledge to regularize the learning of such a sequence-to-sequence model. 
We collect a set of challenging aerial videos and compare the proposed method to the alternative algorithms. Empirical studies suggest that our sequential model achieves significant improvement over alternative image-based methods and the knowledge-rich regularization further boosts our system performance. Our codes and dataset could be found at  https://github.com/Paul-LiPu/DeepVideoHomography
\vspace{5mm}
\end{abstract}

\section{Introduction}

 Estimating planar homographic transformations between consecutive video frames is a fundamental image task and is an essential part of many multimedia applications, including video management, robot navigation, content-based video retrieval, intelligent drones, self-driving vehicles, etc. The estimated homography matrix, for example, can be used for image stitching~\cite{brown2007automatic}, image completion~\cite{huang2014image}, monocular SLAM~\cite{mur2015orb}, camera calibration~\cite{zhang2000flexible}, 3d scene reconstruction~\cite{zhang19963d}, and camera tracking~\cite{simon2000markerless}. The matrix can also be used to approximate camera poses in aerial videos where facing-downward cameras are positioned far away from the ground and objects on the ground could be reasonably assumed to be from the same planar.

\begin{figure}[http!]
  \includegraphics[width=\columnwidth]{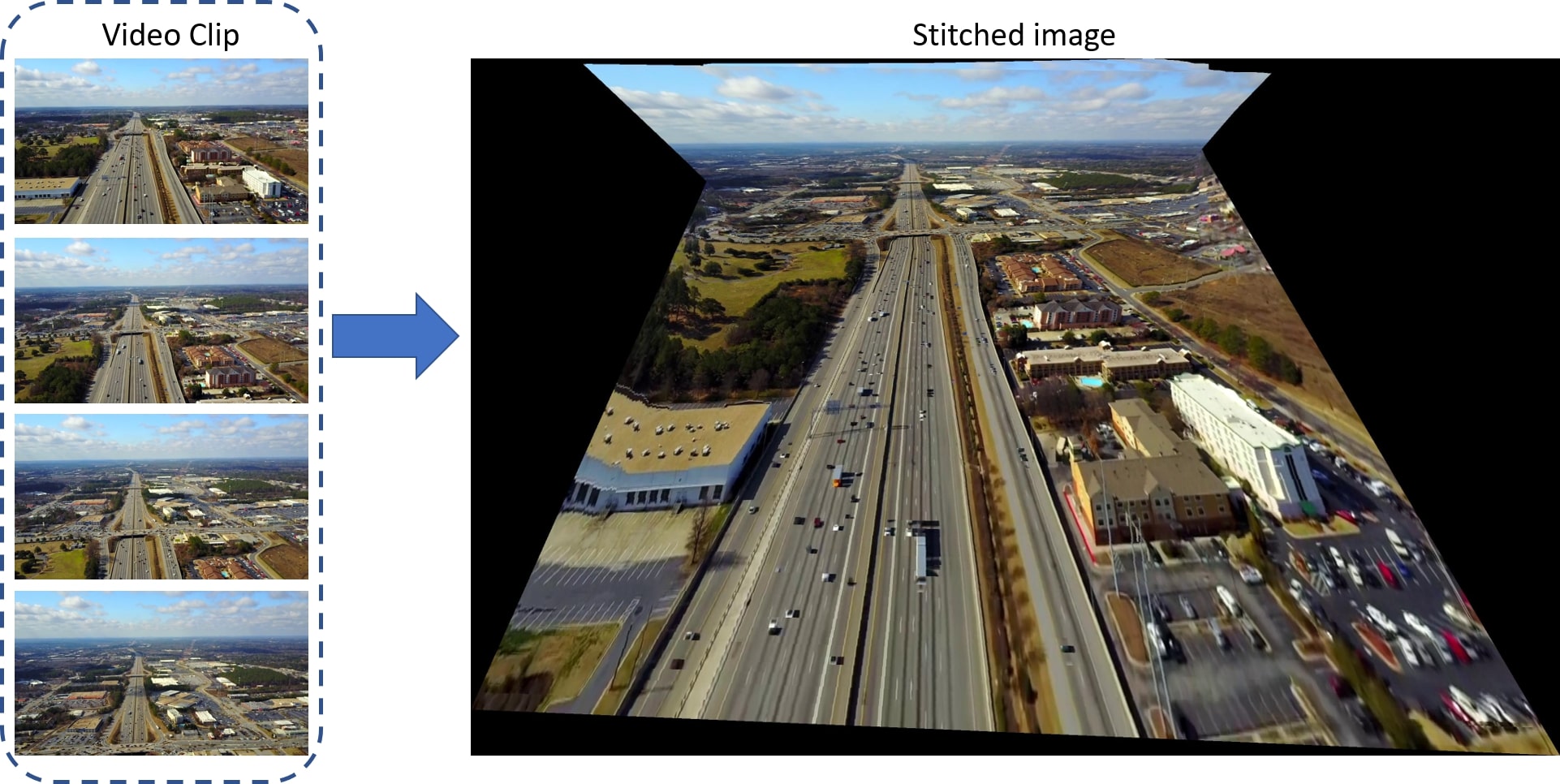}
  \caption{Planar homography estimation and image stitching for aerial videos. Left: a sequence of video frames; right: the scene image stitched by the estimated homographic transformations.}
  \label{fig:fig1}
\end{figure}

 In the past literature, classical estimators usually build feature-level~\cite{lowe2004distinctive, rublee2011orb} or pixel-level~\cite{lucas1981iterative, baker2004lucas, evangelidis2008parametric} correspondences between camera views and employ perspective geometry to recover their planar transformations. Each feature or pixel is represented as a feature descriptor (e.g., histograms) and is matched across camera views. Robust matching methods, e.g., RANSAC~\cite{fischler1981random}, might be used for eliminating incorrect or noisy correspondences. The feature-based methods are robust to some extent because the detected feature points and feature descriptors are invariant to rotation and scaling, and the RANSAC algorithm can potentially prune outlier correspondences. However, such a stage-wise pipeline involves many hand-crafted parameters, including thresholds for detecting feature points, choices of designing feature descriptors, and algorithms for matching feature descriptors. Those hyper-parameters could be tricky to calibrate in real-world multimedia applications.

A recent research stream aims to take advantage of the representative power of deep neural networks to develop end-to-end learning-based estimators. Detone et al.~\cite{detone2016deep} propose to regress planar homography parameters from a pair of input images. A multi-layer convolutional neural network (CNN) is trained using synthetic data which includes multiple pairs of images transformed with randomly generated homography matrices. 
Nguyen et al.~\cite{nguyen2018unsupervised} introduce an unsupervised method that does not need labels of homography matrices to train the deep networks, as reviewed in Section~\ref{sect:one}. Such an unsupervised method has better adaptability and stronger performance than the supervised one~\cite{detone2016deep} but may suffer from severe over-fitting issues. 

To address the above concerns, in this work, we develop a knowledge-rich sequence-to-sequence model to regress homography parameters for aerial videos. 
Figure~\ref{fig:fig1} shows a typical result of the proposed method, which takes a sequence of video frames as inputs and generates a stitched image by a sequence of estimated homography matrices between consecutive video frames. For each pair of input images, our model employs a deep neural network to extract the feature representation of video frame pairs, regresses homography parameters, and warps one input image into the other image. A Long Short-Term Memory (LSTM) model is also employed to directly incorporate the temporal dependencies while estimating the sequence of homography matrices. Our method does not require any annotations and can be trained from raw video frames using standard gradient-based methods. 

We also introduce a set of prior knowledge to regularize the learning of the proposed sequential estimator. These knowledge explicitly impose consistency constraints that should be satisfied while estimating the sequence of homography matrices from the input video sequence. One of the key observations is that for most aerial videos, the camera motion is smooth over time. The homographic transformations between consecutive frames are not arbitrarily different from each other and should be estimated in a coordinated fashion. This leads to a set of 
temporal knowledge
Moreover, we might extract multiple image regions of different resolutions from video frames and estimate the homographic transformations from these image regions. Similarly the estimations cross image regions or scales in the same video frames should be consistent with each other, which leads to set of 
spatial and scale 
knowledge. Incorporating these different types of knowledge is capable of suppressing over-fitting issues while learning the deep model from raw videos. 

\begin{figure*}[http!]
\centering
  \includegraphics[width=0.9\textwidth]{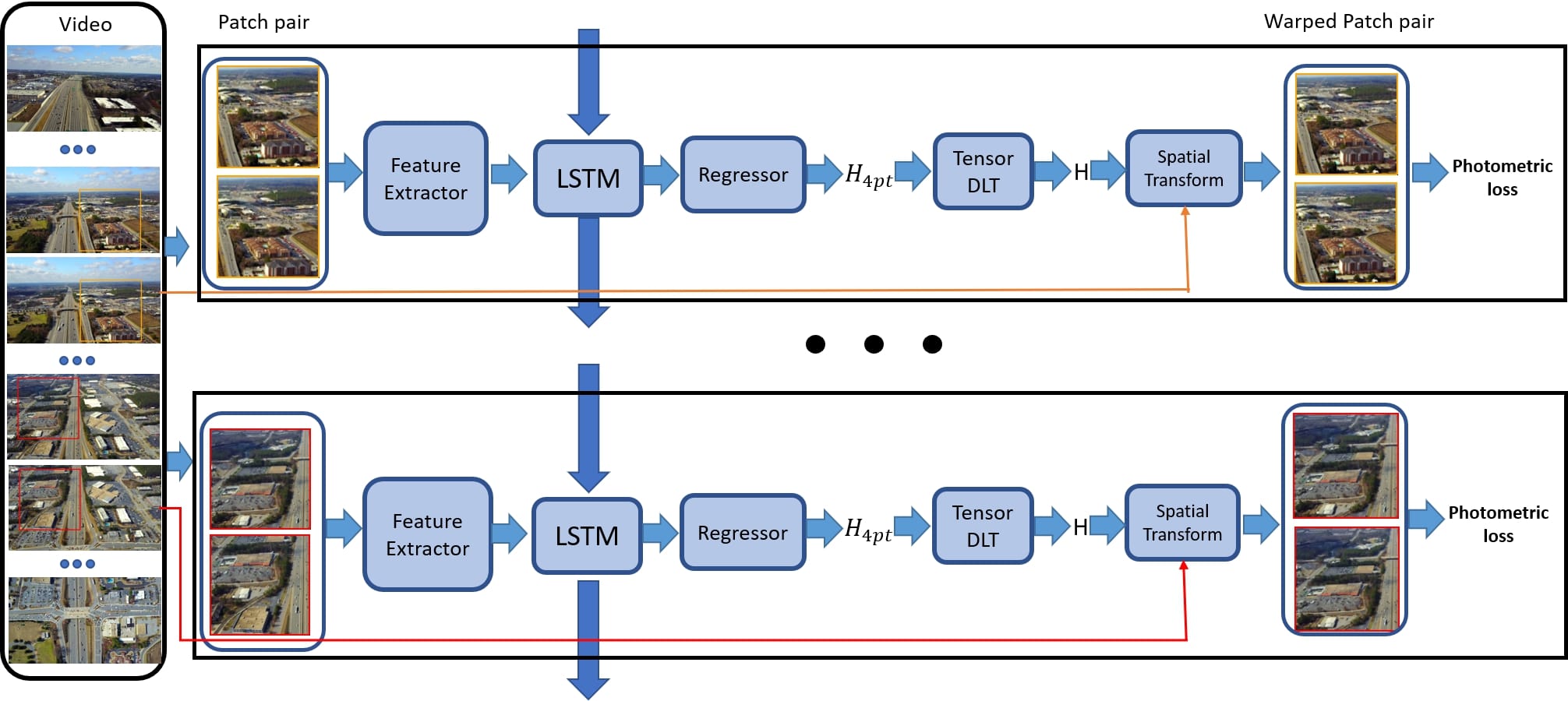}
  \caption{Network architecture of the proposed sequence-to-sequence model. The network takes as inputs one reference image and one target image. \textbf{Feature extractor}: the same VGG-like convolutional layers as the implementation in \cite{nguyen2018unsupervised}, which converts two input patches to a one-dimensional vector of features. \textbf{LSTM}: LSTM layers that integrate the extracted 1D features over time. \textbf{Regressor}: a fully connected layer that aims to regress the offset matrix {$\textbf{H}_{4pt}$}, which includes 4-corner offsets between the two input patches. \textbf{Tensor DLT}: a differentiable layer  \cite{nguyen2018unsupervised} that calculates homography matrix from the patch corner coordinates and offset matrix. \textbf{Spatial Transform}: spatial transform layers that warp one input image into the other one using the estimated homography matrix. \textbf{Photometric loss}: l1 loss between patches in reference frame and warped target frame.}
  \label{fig:fig2}
\end{figure*}

\section{Relationships to Previous Works}
This work is closely related to three research streams in the areas of multimedia retrieval and video analysis. 

\textbf{Geometry-based Homography Estimation}
As aforementioned, classical homography estimators need to match features of interest or pixels across camera views and employ perspective geometry to recover the transformation parameters. 
The feature-based methods~\cite{brown2007automatic, li2015dual, zhao2016accurate, hua2019feature} require to fine-tune many hand-crafted parameters for different scenes and may fail when there are few feature points. The pixel-based algorithms~\cite{chen2018generalized, kang2019combining} are usually of high complexity, which makes them unsuitable for time-critical video applications. For example, the classical estimator ECC \cite{evangelidis2008parametric} method takes around 600ms while the RANSAC~\cite{fischler1981random} with SIFT~\cite{lowe2004distinctive} method,  takes 45ms to process a pair of input images of 320 by 180 pixels on our experiment computer. 
In this work, we present an end-to-end network that is robust to different situations and can be implemented in high-performance parallel computing platforms (e.g., Graphical Processing Units) to boost system efficiency. It only takes around 15ms for our network to process a pair of input images (experiment settings introduced later). 

\textbf{Learning-based Homography Estimation} The recently developed network-based methods~\cite{detone2016deep, zeng2018rethinking, erlik2017homography, nguyen2018unsupervised} present a promising direction to estimating homographic transformations.
To train these networks, it is a common practice to generate a homography matrix and apply it to warp an image to be another image, which leads to a pair of training images with known homography parameters. These deep models can be effectively trained from the generated samples without human in the loop. The major concern is however the known over-fitting issue, i.e., a model might overly fit the synthetic samples so that it can hardly generalize to unseen realistic image pairs. In practice, the training scenarios are often very different from the testing scenarios, which makes the training situation even worse. 
To address the above concerns, we present a sequence-to-sequence LSTM model to estimate a sequence of homography matrices in batches and employ a set of prior knowledge to regularize the learning of such a deep model. Our method can be effectively trained from raw video sequences rather than static images, while satisfying various temporal, spatial and scale knowledge.  

\textbf{Sequence-to-Sequence Models} Recurrent neural networks can process sequential inputs and propagate internal states of certain inputs to their sequential neighbors. 
They have been successfully used to model different sequential data, including language\cite{mikolov2010recurrent}, and skeleton-base action signal~\cite{du2015hierarchical}, etc. Among different RNN models, Long Short-Term Memory(LSTM) is a popular choice for its ability to capture long-term dependencies along with short-term memory on sequential data~\cite{graves2013hybrid, sak2014long, li2015constructing, ma-hovy-2016-end}, including video data~\cite{yao2017boosting}. For aerial videos that are mostly undergoing smooth camera motions, there are strong correlations among the planar transformations on video frames in the same sequence. In this work, we extend the previous network-based estimators using LSTM techniques, which is the first piece of work in its catalog.

The \textbf{Contributions} of this work are two folds. Firstly, we reformulate the homography estimation of aerial videos to be a sequence-to-sequence task and develop a LSTM network to estimate the sequence of homography parameters, which is the first of its catalog in the literature. Secondly, we employ a set of spatial-scale-temporal knowledge to regularize training of the LSTM model and empirically validate its superior performance over alternative methods on challenging aerial videos.

\section{Our Approach}
This section presents the proposed learning-based method for estimating homography parameters in aerial videos. 
In the rest of this section, we first review the previous deep homography method~\cite{nguyen2018unsupervised} in Section~\ref{sect:one}, and then introduce our deep sequence-to-sequence estimator in Section~\ref{sect:two}. In Section~\ref{sect:three}, we present a set of prior knowledge and discuss how they can be leveraged to guide the learning of our sequential model.

\subsection{Background: Deep Homography}\label{sect:one}
The unsupervised network-based method proposed by Nguyen et al.~\cite{nguyen2018unsupervised} takes a pair of images as inputs.
It first crops a pair of image patches from input images, and then employs a network to regress the transformation parameters between the two input images.
The network includes three major components:
(i) A network backbone is used to regress the offsets between the corner coordinates of the two corresponded image patches. This module outputs a 2 by 4 offset matrix $H_{4pt}$. (ii) A Tensor Direct Linear Transform module~\cite{nguyen2018unsupervised} is introduced to calculate the corresponding 3 by 3 homography matrix $\mathrm{H}$ from the offset matrix $H_{4pt}$ and corner coordinates of the image patches. (iii) A Spatial Transformation Layer is used to register one image on the other using $\mathrm{H}$.  With the registered image pair, a pixel-wise photometric loss is calculated to guide the training of such a deep model. The network is fully differentiable and can be trained using standard gradient-based methods. The detailed network architecture is showed in Table~\ref{tab:table_network}.%

This deep homography network~\cite{nguyen2018unsupervised} is trained on a collection of raw aerial images and clearly outperforms the previous supervised methods~\cite{detone2016deep} over images with illumination noises. A significance of this method is that neither human annotations nor synthetic data~\cite{detone2016deep} is needed for the training procedure. This unsupervised method is designed for dealing with pairs of static images.  In this work, we extend it by introducing a deep sequence-to-sequence model to directly deal with aerial videos.

\subsection{Deep Sequence-to-Sequence Estimator} \label{sect:two}

Our sequence-to-sequence network takes a sequence of video frames as inputs and aims to output a homography matrix between every pair of consecutive video frames. Figure \ref{fig:fig2} summarizes the sketch of the proposed network. For every two consecutive video frames, we sample multiple image patches from each frame, and employ a feature extractor to convert a pair of patches into a 1024-dim feature vector. 
We set the size of the input image patch to be 128 by 128 pixels in this work.
A one-hidden-layer LSTM module is used to recurrently integrate the features of a patch pair and that of its preceding video frames. Then a regression layer is integrated to map the 1024-dim features from the LSTM layer into the corner offset matrices $\textbf{H}_{4pt}$. Similar to ~\cite{nguyen2018unsupervised}, a Tensor Direct Linear Transformation (DLT) module is employed to estimate the homography \textbf{H} and a Spatial Transform Layer is used to register the two input images together. 

We employ a pixel-wise photometric loss function defined over the two registered input images~\cite{nguyen2018unsupervised}. 
Denote \textbf{I} as the input video sequence, $I$ and $J$ as two image patches sampled from two consecutive video frames.
Let $I(\textbf{x})$ return the image intensity at the homogeneous coordinate $\mathrm{\textbf{x}\ =\ }{\mathrm{[u,\ v,\ 1]}}^T$, ${\mathrm{\textbf{H}}}_{{\mathrm{i}},{\mathrm{j}}}$ denote the estimated homography matrix between the image patches $I$ and $J$. 
The pixel-wise loss function is defined as: 
\begin{equation}  \label{eq:eq_loss}
    L(\textbf{I})=\sum_{I} \sum_{J} \sum_{\textbf{x} \in I}  \left \| I (\textbf{x}) - J (\hat{\mathrm{\textbf{H}}}_{i,j}\cdot \textbf{x})   \right \|_1
\end{equation}
where $\cdot$ indicates the matrix-vector multiplication between a homography matrix and a coordinate vector.  

It is noteworthy that the above loss function is fully unsupervised as it does not require annotations of homography matrices between video frames. The homography matrix is directly regressed from the input image pair and is used to warping one image into the other. In testing, however, we could output the estimated homography matrix for each pair of images.
Our LSTM model explicitly incorporates the frame-to-frame correlations while estimating the sequence of homography matrices. This is crucial because, for example, two consecutive homography matrices over time might not be arbitrarily different but are closely dependent to each other. 

\begin{table}[t]
\centering
\caption{Network structure}
  \label{tab:table_network}
\begin{tabular}{cccc}
Type                   & Filters              & Size                 & Output               \\ \hline
Convolutional          & 64                   & 3x3                  & 128x128              \\
Convolutional          & 64                   & 3x3                  & 128x128              \\
Max Pooling            &                      & 2x2                  & 64x64                \\
Convolutional          & 64                   & 3x3                  & 64x64                \\
Convolutional          & 64                   & 3x3                  & 64x64                \\
Max Pooling            &                      & 2x2                  & 32x32                \\
Convolutional          & 128                  & 3x3                  & 32x32                \\
Convolutional          & 128                  & 3x3                  & 32x32                \\
Max Pooling            &                      & 2x2                  & 16x16                \\
Convolutional          & 128                  & 3x3                  & 16x16                \\
Convolutional          & 128                  & 3x3                  & 16x16                \\
Fully-connected        &                      & 1024                 &                      \\
LSTM                   &                      & 1024                 &                      \\
Fully-connected        &                      & 8                    &                      \\
Tensor DLT             & \multicolumn{1}{l}{} & \multicolumn{1}{l}{} & \multicolumn{1}{l}{} \\
Spatial Transformation & \multicolumn{1}{l}{} & \multicolumn{1}{l}{} & \multicolumn{1}{l}{}
\end{tabular}
\end{table}

\subsection{Knowledge-rich Regularization}
\label{sect:three}
We develop a set of regularization items to enforce knowledge-compatible consistency among the sequence of homography matrices estimated by the proposed sequence-to-sequence model.
The previous network-based methods~\cite{nguyen2018unsupervised} that work on image pairs have severe overfitting problems while applying over aerial videos. Similarly, we employ a LSTM model that bears a higher model complexity than the network in~\cite{nguyen2018unsupervised} and is more prone to encountering overfitting issues. We therefore derive a set of knowledge in temporal, spatial and scale spaces, respectively, and use them to regularize the loss function. These knowledge play important roles in preventing overfitting issues while training our model.

\textbf{Spatial regularization}
Our method randomly samples multiple image patches from each video frame pair to form training samples. Each image patch is of 128 by 128 pixels. Multiple pairs of image patches could be generated from a single pair of video frames. This sampling strategy, like the data augmentation methods~\cite{cubuk2019autoaugment}, can largely increase the size of training samples and boost system performance without access to extra aerial videos.  

We introduce consistency constraints between the different patch pairs drawn from the same video frame pair. It is reasonably expected that these patch pairs would lead to the exactly same homography matrix. These consistency constraints are specified in the image space and are referred to as spatial knowledge in this work. 
Let $a, b$ index the homography matrices estimated from two pairs of video frames, $H_{a,t,t+1}$ the a-th homography matrix estimated for the frames t and t+1. We have the following the regularization item to encode the spatial knowledge:
\begin{equation} \label{eq:eq_reg_p}
R_{p}(\textbf{I}) = \sum_{t} \sum_{a \neq b} \| H_{a, t, t+1}-H_{b, t, t+1} \| _1
\end{equation}
which calculates the absolute differences between the estimated homography matrices.  
The above $\ell$-1 term is empirically more robust than other norms (e.g., $\ell$-2 norm) especially while dealing with outliers or noises.

\begin{figure*}[t]
\centering
  \includegraphics[width=\textwidth]{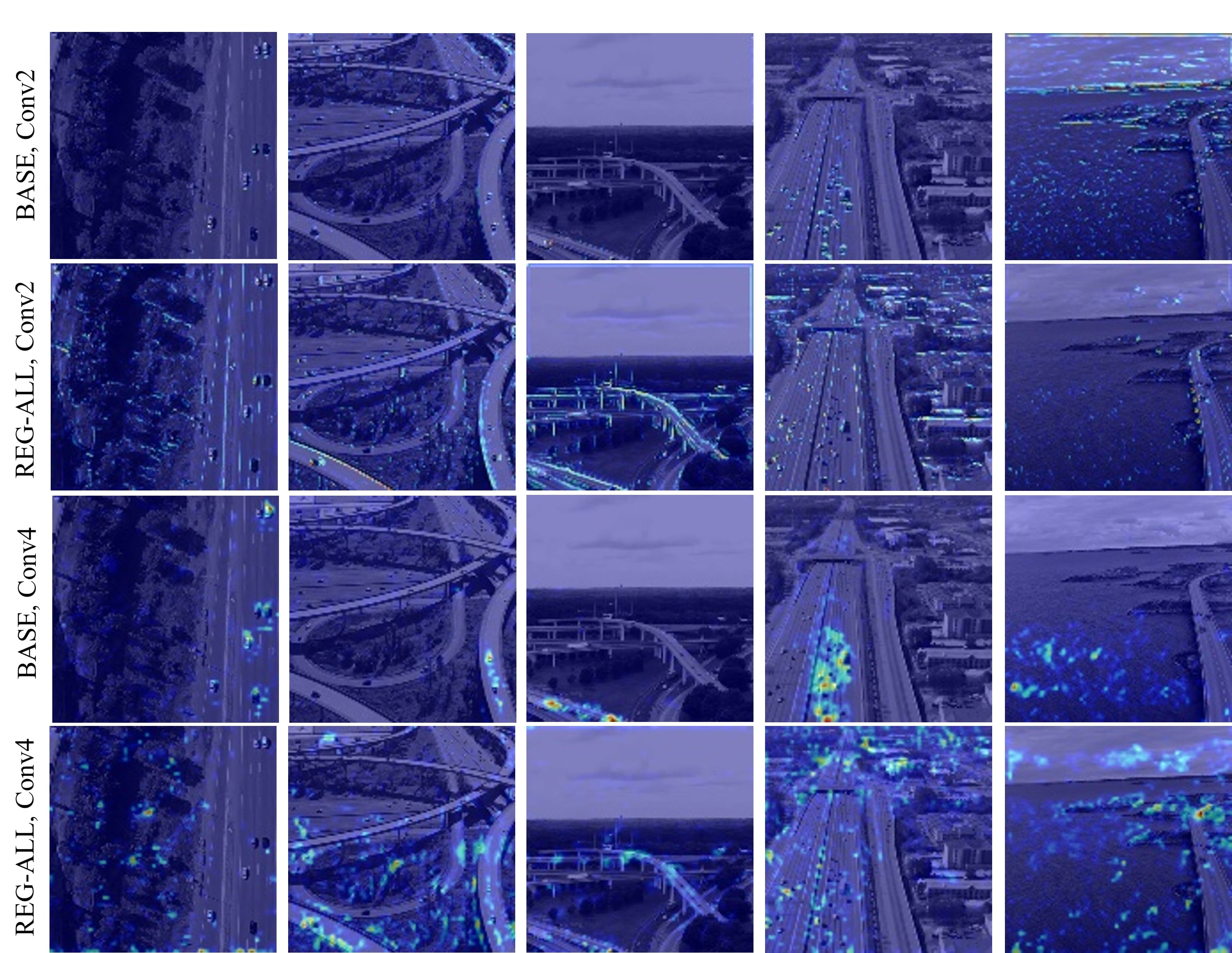}
  \caption{Visualization of neuron activation for the two trained networks: \textbf{BASE} (the first and third rows) and \textbf{REG-ALL} (the second and fourth rows) over five input images (one per column). Rows 1-2: activation on Conv2 layer;  Row 3-4: activation on Conv4 layer. 
  Pixels with warmer colors indicate higher activation for estimating the target homography parameters. 
  }
  \label{fig:fig3} 
\end{figure*}

\textbf{Scale regularization}
We employ multiple patch resolutions while drawing image patches from video frames, which results in a multi-scale representation of the input aerial video. Image patches of different scales can provide rich representations of the input image, as described by the image scaling theory~\cite{wu2008information}. 
This multi-scale information is critical for regressing homography parameters as well. In this work, we employ two scales while sampling image patches from the input videos. We first sample image patches of 128 by 128 pixels and divide each patch into four non-overlapping patches of 64 by 64 pixels. Once drawn, each image patch is resized to be 128 by 128 pixels and fed into the deep LSTM model. 
Let $m,n$ index two homography matrices so that the m-th matrix is estimated from a parent patch and the n-th matrix is from the related children patches. Let $H_{m, t, t+1}$ the m-th homography matrix. We have the following equation to encode the scale consistency: 
\begin{equation}\label{eq:eq_reg_s}
R_{s}(\textbf{I}) =\sum_{t} \sum_{<m,n>} \| H_{m, t, t+1}-H_{n,t, t+1} \| _1
\end{equation}
which calculates the cross-scale differences between the estimated matrices. Such a $\ell$-1 regularization is imposed across scales while the Eq~\eqref{eq:eq_reg_p} only involves image patches at the same scale.

\textbf{Temporal regularization}
We employ two types of temporal knowledge to regularize the learning of the proposed LSTM model. The first type of temporal knowledge is derived from the fact that consecutive video frames tend to undergo similar transformation in aerial videos. Taking an aerial video sequence of 24 FPS for an instance, the duration of three video frames is as short as 83 ms. Therefore, we introduce a regularization item to minimize the absolute differences between the estimated homography parameters for consecutive video frame pairs. 
Let $k, l$ index the homography matrices estimated for consecutive video frames, $ H_{k, t, t+1}$ indicate the k-th matrix estimated for the frames t and t+1. We have a regularization term, 
\begin{equation}\label{eq:eq_reg_t1}
R_{t1}(\textbf{I}) = \sum_{t} \sum_{<k,l>} \| H_{k, t, t+1}-H_{l, t+1, t+2} \| _1
\end{equation}
Minimizing the above term will encourage the predictions of homography parameters not to have dramatic changes over time and thus
encode the smoothness constraint over camera motion in aerial videos. 

The second type of temporal knowledge is used to encode the internal consistency between the homographic transformations in a sequence. For any given pixel \textbf{x} at time $t$,  to obtain its projection $\hat{\textbf{x}}$ on the frame at time $s$, where $s>t$, we might sequentially apply the homography matrices from time t and s. Let $\mathrm{\textbf{H}}_{[t,s]}=\mathrm{\textbf{H}}_{{\mathrm{s-1}},{\mathrm{s}}} \cdot \ldots  \mathrm{\textbf{H}}_{{\mathrm{t+1}},{\mathrm{t}+2}} \cdot  \mathrm{\textbf{H}}_{{\mathrm{t}},{\mathrm{t}+1}}$, and we have, $\hat{\textbf{x}}= \mathrm{\textbf{H}}_{[t,s]} \cdot x$. The intensity values of $\textbf{I}_{t}(\textbf{x})$ and $\textbf{I}_{s}(\hat{\textbf{x}})$ are expected to be 
similar if the sequence of homography estimations are accurate enough. This leads to another temporal consistency constraint. Let K denote the length of the sequence used for our LSTM model, we have the following equation to encode such a temporal constraint, 
\begin{equation}\label{eq:eq_reg_t2}
    R_{t2}(\textbf{I})= \sum_{t} \sum_{s=t+2}^{t+K-1}  \sum_{\textbf{x} \in I_t}  \left \| I_{t} (\textbf{x}) - I_{s}(\mathrm{\textbf{H}}_{[t,s]} \cdot \textbf{x})   \right \|_1
\end{equation}
The above equation is defined over an episode of $K$ video frames and is complementary to the cross-frame loss function in Eq.~\eqref{eq:eq_loss}. Note that we choose to use an episode of $K=16$ video frames in this work. Using a higher order for the video episodes might capture stronger temporal constraints but will also increase the computational complexity. 

In summary, our LSTM aims to minimize the following loss function with knowledge-rich regularization terms. 
\begin{equation}\label{eq:eq_loss_all}
\mathrm{Loss}=L(\textbf{I})+\lambda_p R_{p}(\textbf{I})+\lambda_s R_{s}(\textbf{I})+\lambda_{t1} R_{t1}(\textbf{I})+\lambda_{t2} R_{t2}(\textbf{I}) 
\end{equation}
where $\lambda_p,\lambda_s,\lambda_{t1}, \lambda_{t2}$ are weighting constants. We set $\lambda_p,\lambda_s, \lambda_{t1}$ to be $1/(KN)$ where $K=9$ is the number of elements in a homography matrix, and $N$ is the number of samples involved in each associated regularization term. Similarly, we set $\lambda_{t2}$ to be $1/N$. Minimizing the various regularization terms is capable of narrowing down the feasible space in optimization and suppressing the effects of over-fitting, which is a major concern for training an unsupervised homography network.

\section{Experiment}
\textbf{Datasets\footnote{The dataset and source codes of this work are available under this link: https://github.com/Paul-LiPu/DeepVideoHomography}} 
We collect a set of 
%
aerial videos for training and testing purposes. The training set includes 141 video clips, each lasting between 20 to 40 seconds. All videos share the same aspect ratio of 16:9 and the frames are resized to 320 by 180 pixels before training. We randomly sample image patches at two scales: 128 by 128 and 64 by 64 pixels. In addition to these training videos, we collect another 22 video clips of 1280 x 720 pixels for testing purpose.

\textbf{Evaluation Metrics}
 We evaluate the proposed methods in both qualitative and quantitative ways. To quantity the estimation results, we reported MACE (Mean Average Corner Error)~\cite{detone2016deep, kang2019combining, erlik2017homography, zeng2018rethinking, wang2019self}.
 For each testing video, we identify multiple landmark points (e.g. corners of buildings) and annotate their image locations every 30 frames. We denote \textbf{M} as the number of testing sequences, $N_i$ as the total number of annotated frames for i-th testing video,${K_t}$ as the number of annotated landmark points between (t-1)-th and t-th annotated frames, $\hat{x_j}^t$ and $x_j^{t}$ as j-th predicted landmark coordinates and ground truth landmark coordinates respectively. The MACE is calculated as:
\begin{equation}
MACE=\frac{1}{\sum_{i=1}^{M}(N_i\mathrm{-}1)}\sum_{m=1}^{M}\sum_{t=2}^{N_i} (\frac{1}{K_t}\sum_{j=1}^{K_t} \| \hat{x_j}^t\mathrm{-} x_j^{t} \| _2 )
\end{equation} 
where $\|\|_2$ is the $\ell-2$ norm of a vector. 
These coordinates are registered to original video frame of 1280 by 720 pixels. The predicted landmark coordinates are obtained from last annotated landmark location and sequential homographic transformations among the 30 frames. 
 
\textbf{Baseline methods}
We compare the proposed sequence-to-sequence model to the original unsupervised method~\cite{nguyen2018unsupervised}. We apply the image-based method~\cite{nguyen2018unsupervised} over every pair of consecutive video frames separately and assemble all the estimated matrices together for evaluation purposes. We also compare to the traditional feature based method that extracts ORB~\cite{rublee2011orb} features and apply RANSAC~\cite{fischler1981random} to get homographic transformations. 
We apply the ORB-RANSAC method over every pair of consecutive video frames. We also evaluate identity matrix as estimated homography and include its results for comparison.

\textbf{Implementation Variants}
We implement seven variants of the proposed method for analyzing the individual contributions of the proposed techniques. (1)
BASE, the previous unsupervised method~\cite{nguyen2018unsupervised} with the loss function in Eq.~\eqref{eq:eq_loss};
(2) REG-P, learning the baseline network BASE with the spatial regularization term Eq.~\eqref{eq:eq_reg_p};  (3) REG-S, learning BASE with the scale regularization Eq.~\eqref{eq:eq_reg_s}; (4) REG-T, learning BASE with the two temporal regularization terms Eq.~\eqref{eq:eq_reg_t1} and Eq.~\eqref{eq:eq_reg_t2}; (5) REG-ALL, learning BASE with all regularization terms; (6) LSTM, the proposed LSTM model with the loss function in Eq.~\eqref{eq:eq_loss} not regularized by any knowledge; (7) LSTM-REG-ALL, the LSTM model that employs all the three types of knowledge. 

We train the BASE network for 300k iterations with a mini-batch size of 64 (each sample contains a patch pair from a frame pair). The initial learning rate is $0.001$ and learning rate is multiplied by 0.1 every 100k iterations. The networks REG-T, REG-S, REG-P
and REG-ALL are finetuned from BASE model with a batch size of $32$ for 90k iterations. The LSTM networks are finetuned for 90k iterations with batch size of 8. The learning rate is initially $0.001$ and decayed by factor 0.1 every 30k iterations for those experiments.
All networks are implemented using Pytorch~\cite{NEURIPS2019_9015} and are initialized as introduced in~\cite{he2015delving}. The Adam optimizer~\cite{kingma2014adam} with default parameters is used for optimization. 


\textbf{Efficiency} We implemented the proposed algorithms using a computer with Intel Core i5-4690 CPU and NVIDIA GTX-1080ti GPU. It takes 13.4 ms for LSTM-REG-ALL to process a video frame. The processing time includes image reading and preprocessing, and is averaged over all testing frames. The average run time of BASE, REG-T, REG-P, REG-S, REG-ALL is about 13.2 ms per frame. Note that the proposed regularization terms do not affect the network inference nor testing efficiency. 
The LSTM models with additional LSTM layers involve about 0.7\% more computation than other networks. Moreover, the BASE and LSTM models have FLOPS of 1.27G and 1.28G respectively, which are very similar to each other. 

\textbf{Network Visualization}. Once trained the BASE and REG-ALL networks, we first visualize the importance of each pixel for regressing the target homography matrix.
For a testing image pair, we feed it to the network layer by layer, calculate its loss by Eq.~\eqref{eq:eq_loss}, and backward propagate the loss to each layer. We then visualize the neuron activation weighted by gradients over every point in each feature map. 
More details about the visualization method can be found in the previous work~\cite{selvaraju2017grad}. Figure~\ref{fig:fig3} visualizes feature maps from Conv2 (top two rows) and Conv4 (bottom two rows) layers of the networks BASE (Rows 1 and 3) and REG-ALL (Rows 2 and 4), respectively. Five testing images are used for visualization purposes. 
From these comparisons, we can observe that enforcing various knowledge can significantly enhance the local distinctness of the feature maps at different layers. Taking the layer Conv2 and the first image for an example, the highlighted pixels for the BASE model (row-1) spread over the whole image while those for the REG-ALL model (row-2) tend to appear in a few clustered regions. Similar results can be observed for other examples.

\begin{figure}[t]
  \includegraphics[width=\columnwidth]{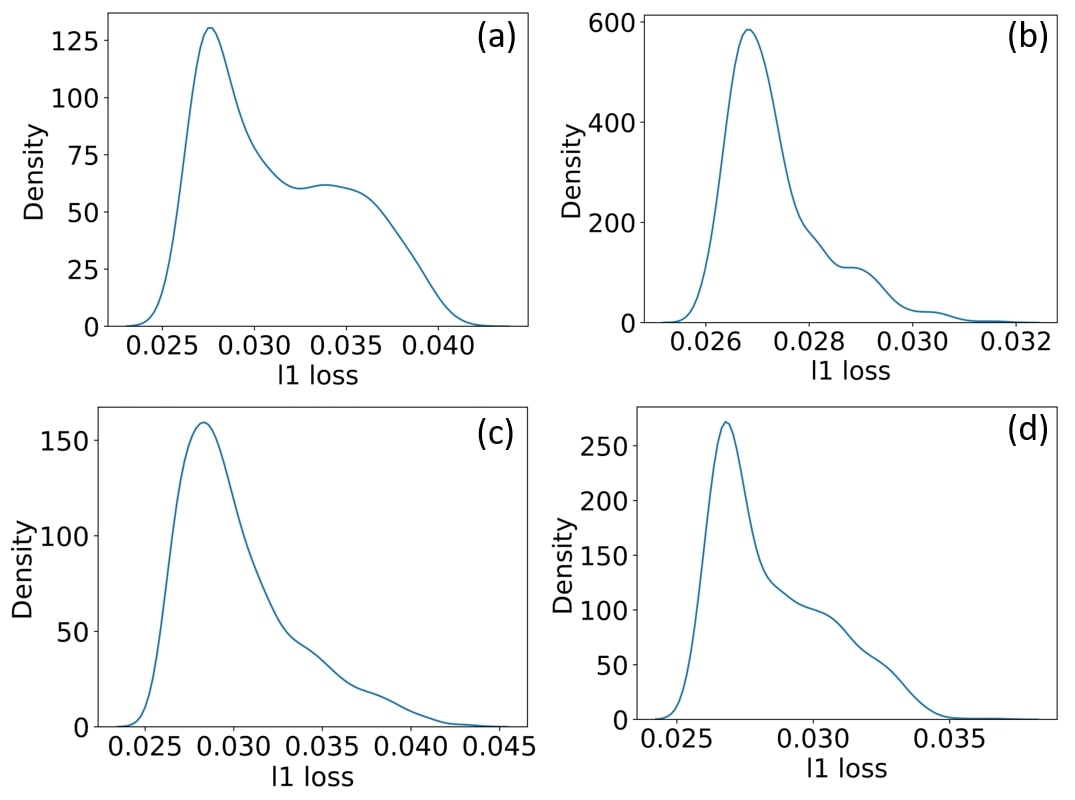}
  \caption{Distributions of photometric loss values over a set of 1000 pairs of image patches sampled from one pair of video frames. 
  (a) \textbf{BASE}; (b) \textbf{REG-T}; (c) \textbf{REG-P}; (d) \textbf{REG-S}. }
  \label{fig:loss}
\end{figure}

\begin{table}[t]
\centering
\caption{Numerical results (MACE) on testing videos. 
}
  \label{tab:table1}
  \begin{tabular}{|c|c|l|p{\columnwidth}}
    \hline
Experiments & MACE \\ \hline
IDENTITY    & 35.69        \\
ORB2+RANSAC & 12.02        \\
BASE        & 13.66        \\
REG-T       & 9.95         \\
REG-P       & 11.57        \\
REG-S      & 11.44        \\
REG-ALL     & 9.16         \\
LSTM        & 14.10        \\
LSTM-REG-ALL  & \textbf{8.77}         \\ \hline
  \end{tabular}  
\end{table}

\begin{figure*}[ht!]
\centering
  \includegraphics[width=0.9\textwidth]{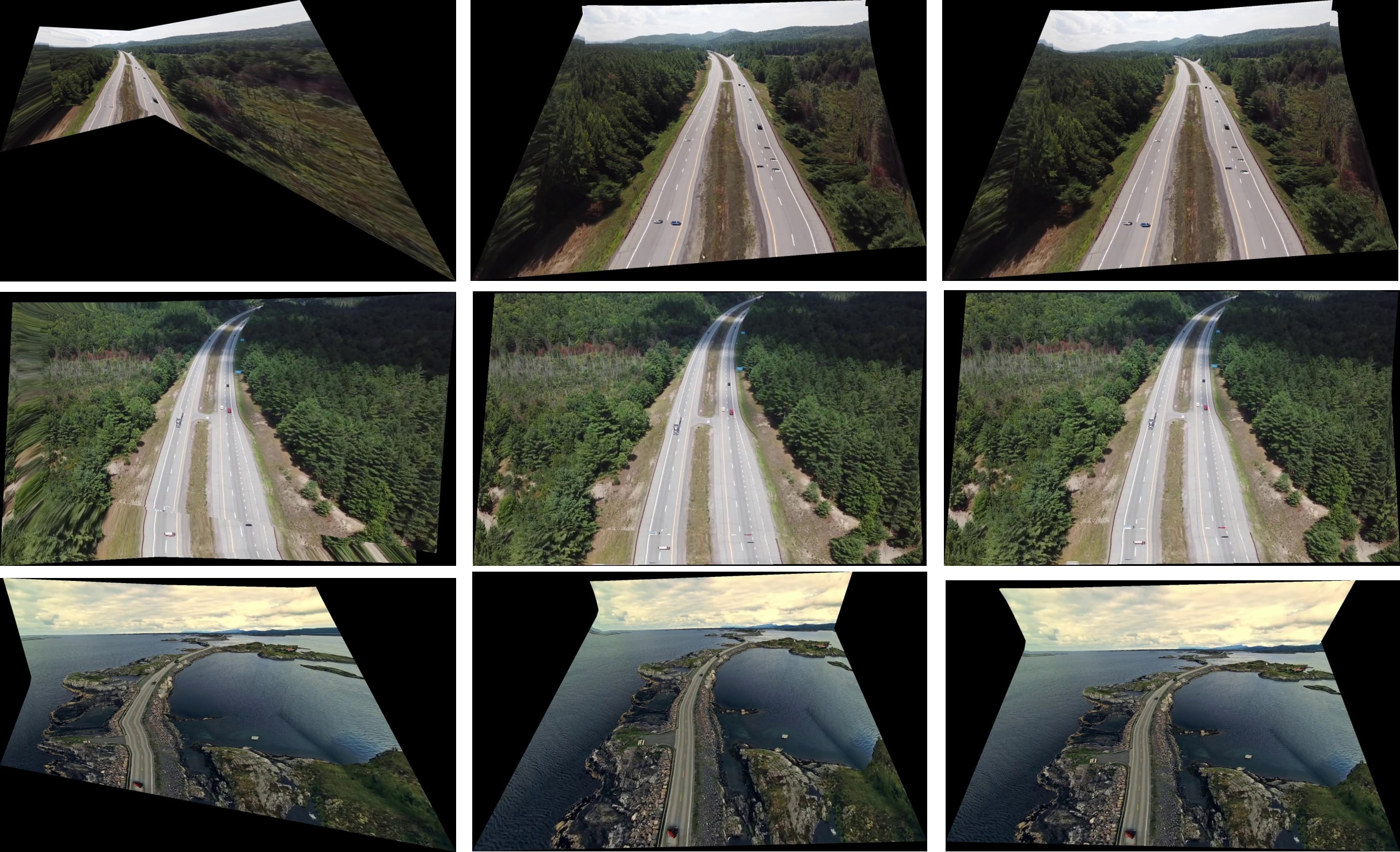}
  \caption{Image stitching results for three testing aerial videos by three methods:  \textbf{BASE} (left column), \textbf{REG-ALL} (middle column) and \textbf{LSTM-REG-All} (right column).}
  \label{fig:stichingresults} 
\end{figure*}

\textbf{Loss Distribution} We examine how the proposed knowledge-rich regularization items affect the photometric loss values over testing images. In this experiment, we randomly select 1000 pairs of image patches from a single pair of video frames, and apply the trained networks BASE, REG-T, REG-P, and
REG-S over each image pair. Figure~\ref{fig:loss} plots the histogram of loss values for these four networks in (a), (b), (c) and (d), respectively. From the results, we can observe that the proposed spatial-scale-temporal knowledge can effectively reduce the expected loss value comparing to the baseline method.

\textbf{Quantitative result}. 
Table~\ref{tab:table1} reports the numerical results of the proposed methods and baseline methods. 
IDENTITY uses identity matrix as estimated homography matrix, while ORB2+RANSAC extracts ORB2 features and applies RANSAC to estimate the homography matrix. BASE, REG-T, REG-P, REG-S, REG-ALL, LSTM, and LSTM-REG-ALL predict the homography parameters by our trained deep models. 
We have two major observations. Firstly, the proposed spatial knowledge (REG-P), temporal knowledge (REG-T) and scale knowledge (REG-S) can significantly boost the unsupervised method BASE with good margins. These comparisons clearly demonstrate the effectiveness of the proposed knowledge-compatible consistency constraints. Secondly, the proposed LSTM model achieve the best performance among all baseline methods and implementation variants of our methods. Notably, the LSTM-REG-ALL model that employs spatial, temporal and scale knowledge outperforms the base LSTM model by a margin of 5.33 in terms of MACE.  

\textbf{Qualitative result} Figure \ref{fig:stichingresults} plots three exemplar results of image stitching using the homographic transformations estimated by three models:  
BASE, REG-ALL and LSTM-REG-ALL. For each video sequence, we apply the three models to estimate the sequence of homography matrices, which are then used to register these video frames together. 
For the video in the first row, both REG-ALL and LSTM-REG-ALL can generate descent results whereas the BASE fails to work. 
For the second video, although all three models work well, the LSTM model generated less distorted pixels in the left side of the stitched image. 
For the third video sequence, the BASE and REG-ALL do not recover the geometry correctly whereas the proposed method can correctly reconstruct the scene structure. These comparisons clearly validate the effectiveness of the proposed sequence-to-sequence model and various knowledge.

\section{Conclusion}
This paper presents a deep sequence-to-sequence method for estimating homographic transformations from raw aerial videos without human annotations. Our method significantly outperforms previous image-based homography estimator on challenging aerial video data. The significance of our results highlight two insights: (i) The proposed sequence-to-sequence model is capable of
leveraging sequential information for video homography estimation; (ii) the various knowledge-compatible constraints can significantly improve system generalization on unseen video data. The proposed techniques have wide potentials in other video tasks, including camera pose estimation, 3D scene reconstruction, etc.  

The proposed method frames a new direction in the area of video registration. Our current method is however limited by the fact that it manually samples image patches from video frames and estimates their transformation parameters. This strategy can be largely improved by a network-based method that learns to select image patches that could best represent the target scenario, e.g., image regions with distinctive features, without object movements, and/or without repetitive patterns. We will continue to investigate this direction in the future. 


\bibliographystyle{IEEEtran}
\bibliography{homography}

\begin{thebibliography}{10}
\providecommand{\url}[1]{#1}
\csname url@samestyle\endcsname
\providecommand{\newblock}{\relax}
\providecommand{\bibinfo}[2]{#2}
\providecommand{\BIBentrySTDinterwordspacing}{\spaceskip=0pt\relax}
\providecommand{\BIBentryALTinterwordstretchfactor}{4}
\providecommand{\BIBentryALTinterwordspacing}{\spaceskip=\fontdimen2\font plus
\BIBentryALTinterwordstretchfactor\fontdimen3\font minus
  \fontdimen4\font\relax}
\providecommand{\BIBforeignlanguage}[2]{{%
\expandafter\ifx\csname l@#1\endcsname\relax
\typeout{** WARNING: IEEEtran.bst: No hyphenation pattern has been}%
\typeout{** loaded for the language `#1'. Using the pattern for}%
\typeout{** the default language instead.}%
\else
\language=\csname l@#1\endcsname
\fi
#2}}
\providecommand{\BIBdecl}{\relax}
\BIBdecl

\bibitem{brown2007automatic}
M.~Brown and D.~G. Lowe, ``Automatic panoramic image stitching using invariant
  features,'' \emph{International journal of computer vision}, vol.~74, no.~1,
  pp. 59--73, 2007.

\bibitem{huang2014image}
J.-B. Huang, S.~B. Kang, N.~Ahuja, and J.~Kopf, ``Image completion using planar
  structure guidance,'' \emph{ACM Transactions on graphics (TOG)}, vol.~33,
  no.~4, p. 129, 2014.

\bibitem{mur2015orb}
R.~Mur-Artal, J.~M.~M. Montiel, and J.~D. Tardos, ``Orb-slam: a versatile and
  accurate monocular slam system,'' \emph{IEEE transactions on robotics},
  vol.~31, no.~5, pp. 1147--1163, 2015.

\bibitem{zhang2000flexible}
Z.~Zhang, ``A flexible new technique for camera calibration,'' \emph{IEEE
  Transactions on pattern analysis and machine intelligence}, vol.~22, 2000.

\bibitem{zhang19963d}
Z.~Zhang and A.~R. Hanson, ``3d reconstruction based on homography mapping,''
  \emph{Proc. ARPA96}, pp. 1007--1012, 1996.

\bibitem{simon2000markerless}
G.~Simon, A.~W. Fitzgibbon, and A.~Zisserman, ``Markerless tracking using
  planar structures in the scene,'' in \emph{Proceedings IEEE and ACM
  International Symposium on Augmented Reality (ISAR 2000)}.\hskip 1em plus
  0.5em minus 0.4em\relax IEEE, 2000, pp. 120--128.

\bibitem{lowe2004distinctive}
D.~G. Lowe, ``Distinctive image features from scale-invariant keypoints,''
  \emph{International journal of computer vision}, vol.~60, no.~2, pp. 91--110,
  2004.

\bibitem{rublee2011orb}
E.~Rublee, V.~Rabaud, K.~Konolige, and G.~R. Bradski, ``Orb: An efficient
  alternative to sift or surf.'' in \emph{ICCV}, vol.~11, no.~1.\hskip 1em plus
  0.5em minus 0.4em\relax Citeseer, 2011, p.~2.

\bibitem{lucas1981iterative}
B.~D. Lucas, T.~Kanade \emph{et~al.}, ``An iterative image registration
  technique with an application to stereo vision,'' 1981.

\bibitem{baker2004lucas}
S.~Baker and I.~Matthews, ``Lucas-kanade 20 years on: A unifying framework,''
  \emph{International journal of computer vision}, vol.~56, no.~3, pp.
  221--255, 2004.

\bibitem{evangelidis2008parametric}
G.~D. Evangelidis and E.~Z. Psarakis, ``Parametric image alignment using
  enhanced correlation coefficient maximization,'' \emph{IEEE Transactions on
  Pattern Analysis and Machine Intelligence}, vol.~30, no.~10, pp. 1858--1865,
  2008.

\bibitem{fischler1981random}
M.~A. Fischler and R.~C. Bolles, ``Random sample consensus: a paradigm for
  model fitting with applications to image analysis and automated
  cartography,'' \emph{Communications of the ACM}, vol.~24, no.~6, pp.
  381--395, 1981.

\bibitem{detone2016deep}
D.~DeTone, T.~Malisiewicz, and A.~Rabinovich, ``Deep image homography
  estimation,'' \emph{arXiv preprint arXiv:1606.03798}, 2016.

\bibitem{nguyen2018unsupervised}
T.~Nguyen, S.~W. Chen, S.~S. Shivakumar, C.~J. Taylor, and V.~Kumar,
  ``Unsupervised deep homography: A fast and robust homography estimation
  model,'' \emph{IEEE Robotics and Automation Letters}, vol.~3, no.~3, pp.
  2346--2353, 2018.

\bibitem{li2015dual}
S.~Li, L.~Yuan, J.~Sun, and L.~Quan, ``Dual-feature warping-based motion model
  estimation,'' in \emph{Proceedings of the IEEE International Conference on
  Computer Vision}, 2015, pp. 4283--4291.

\bibitem{zhao2016accurate}
C.~Zhao and H.~Zhao, ``Accurate and robust feature-based homography estimation
  using half-sift and feature localization error weighting,'' \emph{Journal of
  Visual Communication and Image Representation}, vol.~40, pp. 288--299, 2016.

\bibitem{hua2019feature}
M.-D. Hua, J.~Trumpf, T.~Hamel, R.~Mahony, and P.~Morin, ``Feature-based
  recursive observer design for homography estimation and its application to
  image stabilization,'' \emph{Asian Journal of Control}, vol.~21, no.~4, pp.
  1443--1458, 2019.

\bibitem{chen2018generalized}
K.~Chen, J.~Tu, J.~Yao, and J.~Li, ``Generalized content-preserving warp:
  Direct photometric alignment beyond color consistency,'' \emph{IEEE Access},
  vol.~6, pp. 69\,835--69\,849, 2018.

\bibitem{kang2019combining}
L.~Kang, Y.~Wei, Y.~Xie, J.~Jiang, and Y.~Guo, ``Combining convolutional neural
  network and photometric refinement for accurate homography estimation,''
  \emph{IEEE Access}, vol.~7, pp. 109\,460--109\,473, 2019.

\bibitem{zeng2018rethinking}
R.~Zeng, S.~Denman, S.~Sridharan, and C.~Fookes, ``Rethinking planar homography
  estimation using perspective fields,'' in \emph{Asian Conference on Computer
  Vision}.\hskip 1em plus 0.5em minus 0.4em\relax Springer, 2018, pp. 571--586.

\bibitem{erlik2017homography}
F.~Erlik~Nowruzi, R.~Laganiere, and N.~Japkowicz, ``Homography estimation from
  image pairs with hierarchical convolutional networks,'' in \emph{Proceedings
  of the IEEE International Conference on Computer Vision}, 2017, pp. 913--920.

\bibitem{mikolov2010recurrent}
T.~Mikolov, M.~Karafi{\'a}t, L.~Burget, J.~{\v{C}}ernock{\`y}, and
  S.~Khudanpur, ``Recurrent neural network based language model,'' in
  \emph{Eleventh annual conference of the international speech communication
  association}, 2010.

\bibitem{du2015hierarchical}
Y.~Du, W.~Wang, and L.~Wang, ``Hierarchical recurrent neural network for
  skeleton based action recognition,'' in \emph{Proceedings of the IEEE
  conference on computer vision and pattern recognition}, 2015, pp. 1110--1118.

\bibitem{graves2013hybrid}
A.~Graves, N.~Jaitly, and A.-r. Mohamed, ``Hybrid speech recognition with deep
  bidirectional lstm,'' in \emph{2013 IEEE workshop on automatic speech
  recognition and understanding}.\hskip 1em plus 0.5em minus 0.4em\relax IEEE,
  2013, pp. 273--278.

\bibitem{sak2014long}
H.~Sak, A.~Senior, and F.~Beaufays, ``Long short-term memory recurrent neural
  network architectures for large scale acoustic modeling,'' in \emph{Fifteenth
  annual conference of the international speech communication association},
  2014.

\bibitem{li2015constructing}
X.~Li and X.~Wu, ``Constructing long short-term memory based deep recurrent
  neural networks for large vocabulary speech recognition,'' in \emph{2015 IEEE
  International Conference on Acoustics, Speech and Signal Processing
  (ICASSP)}.\hskip 1em plus 0.5em minus 0.4em\relax IEEE, 2015, pp. 4520--4524.

\bibitem{ma-hovy-2016-end}
\BIBentryALTinterwordspacing
X.~Ma and E.~Hovy, ``End-to-end sequence labeling via bi-directional
  {LSTM}-{CNN}s-{CRF},'' in \emph{Proceedings of the 54th Annual Meeting of the
  Association for Computational Linguistics (Volume 1: Long Papers)}.\hskip 1em
  plus 0.5em minus 0.4em\relax Berlin, Germany: Association for Computational
  Linguistics, Aug. 2016, pp. 1064--1074. [Online]. Available:
  \url{https://www.aclweb.org/anthology/P16-1101}
\BIBentrySTDinterwordspacing

\bibitem{yao2017boosting}
T.~Yao, Y.~Pan, Y.~Li, Z.~Qiu, and T.~Mei, ``Boosting image captioning with
  attributes,'' in \emph{Proceedings of the IEEE International Conference on
  Computer Vision}, 2017, pp. 4894--4902.

\bibitem{cubuk2019autoaugment}
E.~D. Cubuk, B.~Zoph, D.~Mane, V.~Vasudevan, and Q.~V. Le, ``Autoaugment:
  Learning augmentation strategies from data,'' in \emph{Proceedings of the
  IEEE conference on computer vision and pattern recognition}, 2019, pp.
  113--123.

\bibitem{wu2008information}
Y.~N. Wu, C.-E. Guo, and S.-C. Zhu, ``From information scaling of natural
  images to regimes of statistical models,'' \emph{Quarterly of Applied
  Mathematics}, pp. 81--122, 2008.

\bibitem{wang2019self}
C.~Wang, X.~Wang, X.~Bai, Y.~Liu, and J.~Zhou, ``Self-supervised deep
  homography estimation with invertibility constraints,'' \emph{Pattern
  Recognition Letters}, vol. 128, pp. 355--360, 2019.

\bibitem{NEURIPS2019_9015}
\BIBentryALTinterwordspacing
A.~Paszke, S.~Gross, F.~Massa, A.~Lerer, J.~Bradbury, G.~Chanan, T.~Killeen,
  Z.~Lin, N.~Gimelshein, L.~Antiga, A.~Desmaison, A.~Kopf, E.~Yang, Z.~DeVito,
  M.~Raison, A.~Tejani, S.~Chilamkurthy, B.~Steiner, L.~Fang, J.~Bai, and
  S.~Chintala, ``Pytorch: An imperative style, high-performance deep learning
  library,'' in \emph{Advances in Neural Information Processing Systems 32},
  H.~Wallach, H.~Larochelle, A.~Beygelzimer, F.~d\textquotesingle
  Alch\'{e}-Buc, E.~Fox, and R.~Garnett, Eds.\hskip 1em plus 0.5em minus
  0.4em\relax Curran Associates, Inc., 2019, pp. 8024--8035. [Online].
  Available:
  \url{http://papers.neurips.cc/paper/9015-pytorch-an-imperative-style-high-performance-deep-learning-library.pdf}
\BIBentrySTDinterwordspacing

\bibitem{he2015delving}
K.~He, X.~Zhang, S.~Ren, and J.~Sun, ``Delving deep into rectifiers: Surpassing
  human-level performance on imagenet classification,'' in \emph{Proceedings of
  the IEEE international conference on computer vision}, 2015, pp. 1026--1034.

\bibitem{kingma2014adam}
D.~P. Kingma and J.~Ba, ``Adam: A method for stochastic optimization,''
  \emph{arXiv preprint arXiv:1412.6980}, 2014.

\bibitem{selvaraju2017grad}
R.~R. Selvaraju, M.~Cogswell, A.~Das, R.~Vedantam, D.~Parikh, and D.~Batra,
  ``Grad-cam: Visual explanations from deep networks via gradient-based
  localization,'' in \emph{Proceedings of the IEEE international conference on
  computer vision}, 2017, pp. 618--626.

\end{thebibliography}

\end{document}